\documentclass[fleqn,10pt]{wlscirep}
\usepackage{floatrow}
\newfloatcommand{capbtabbox}{table}[][\FBwidth]

\usepackage{blindtext}
\usepackage{hyperref}
\usepackage{mathtools}

\usepackage{color}
\usepackage{tcolorbox}
\usepackage{amsmath}
\usepackage{amsfonts}
\usepackage{amssymb}
\usepackage{graphicx}
\usepackage{enumitem}
\usepackage{float}
\usepackage{longtable}
\usepackage{algcompatible}
\usepackage{algorithm}
\usepackage{caption}
\usepackage{adjustbox}
\usepackage{array}
\usepackage{rotating} 
\usepackage[font=scriptsize, skip=0cm, labelfont=bf]{caption}
\usepackage{booktabs}
\usepackage[justification=centering]{caption}
\usepackage[title]{appendix}
\usepackage{blindtext}

\usepackage{geometry}

\usepackage[utf8]{inputenc}
\usepackage[english]{babel}

\usepackage{tikz}
\usetikzlibrary{shapes,arrows}

\usepackage{xurl}
\usepackage{caption}
\usepackage{subcaption}

\title{A Machine Learning Analysis of COVID-19 Mental Health Data}

\author[1,*]{Mostafa Rezapour}
\author[1]{Lucas Hansen}
\affil[1]{Department of Mathematics and Statistics, Wake Forest University, NC, U.S.}

\affil[*]{Corresponding author email: rezapom@wfu.edu }


\begin{abstract}
In late December 2019, the novel coronavirus (Sars-Cov-2) and the resulting disease COVID-19 were first identified in Wuhan China.  The disease slipped through containment measures,  with the first known case in the United States being identified on January 20th, 2020. In this paper, we utilize survey data from the Inter-university Consortium for Political and Social Research and apply several statistical and machine learning models and techniques such as Decision Trees, Multinomial Logistic Regression, Naive Bayes, k-Nearest Neighbors, Support Vector Machines, Neural Networks, Random Forests, Gradient Tree Boosting, XGBoost, CatBoost, LightGBM, Synthetic Minority Oversampling, and Chi-Squared Test to analyze the impacts the COVID-19 pandemic has had on the mental health of frontline workers in the United States. Through the interpretation of the many models applied to the mental health survey data, we have concluded that the most important factor in predicting the mental health decline of a frontline worker is the healthcare role the individual is in (Nurse, Emergency Room Staff, Surgeon, etc.), followed by the amount of sleep the individual has had in the last week, the amount of COVID-19 related news an individual has consumed on average in a day, the age of the worker, and the usage of alcohol and cannabis.
\end{abstract}
\begin{document}

\flushbottom
\maketitle
%
%
\thispagestyle{empty}

\section{Introduction}

In late December 2019, the novel coronavirus (Sars-Cov-2) and the resulting disease COVID-19 were first identified in Wuhan China \cite{huang2020clinical}. The disease slipped through containment measures \cite{nishiura2020extent}, with the first known case in the United States being identified on January $20^{th}, 2020$ \cite{harcourt2020severe}. As countries around the world grappled with the new disease, new data and its analysis have heavily influenced policymakers around the globe and vastly transformed our knowledge of the disease and its effects \cite {WHO}. In this paper, we examine the indirect effects of COVID-19 on the mental health of frontline workers in the United States and offer approaches to help frontline workers retain their mental health during the COVID-19 pandemic.

The United States Center for Disease Control and Prevention (CDC) defines an epidemic as \textit{an increase, often sudden, in the number of cases of a disease above what is normally expected in the population in an area}. The CDC then defines a pandemic as \textit{an epidemic that has spread over several countries or continents, usually affecting a large number of people}\cite{centers2018lesson}. Following these guidelines, COVID-19 was officially declared a pandemic on March $11^{th}, 2020$ by the World Health Organization (WHO) and is still considered to be an ongoing pandemic. Prior to COVID-19, the most recent public health crisis was Zika Virus, first isolated in Uganda in 1947 \cite{yadav2016zika}. Other recent major infectious disease events include outbreaks of Ebola in 2014, MERS in 2013, SARS in 2003, and H1N1 Influenza (Swine Flu) in 2009.

As the world is battling the COVID-19 pandemic, one of the most vulnerable groups for mental health problems are frontline workers such as nurses, doctors, and emergency room staff. The risks of being on the front lines of combating the COVID-19 pandemic are not well understood \cite{cabarkapa2020psychological} with even less being known about how to ensure the workers remain mentally healthy. Post-SARS research suggests hospital administration and staff needs to recognize that the impact to the health of frontline workers and frequent changing of infectious disease policy can play major roles in detrimentally impacting frontline workers mental health \cite{tam2004severe}. It is known that survivors of infectious diseases have higher rates of Post-Traumatic Stress Disorder (PTSD) \cite{hong2009posttraumatic}.  It is also known that negative outcomes such as anxiety, burnout, and depression have been reported after outbreaks,  suggesting possible long term effects of being on the front line during health crisis \cite{lancee2008prevalence}. Moreover, new research has indicated an increase in physician suicide rates during the COVID-19 pandemic \cite{laboe2021physician}. Similar to physician suicides, during COVID-19 there has been an increase in the in the symptoms of those suffering from psychiatric disorders \cite{jain2020impact}. Notably in the Republic of Ireland, a study found that during the initial phase of the COVID-19 pandemic, depression became more common \cite{hyland2020anxiety}. 

In this paper, we utilize survey data obtained from the University of Michigan's Inter-university Consortium for Political and Social Research (ICPSR). The data was collected by Deirdre Conroy \cite{ConroyDeirdre,conroy2021effects} from the University of Michigan Department of Psychiatry and Cathy Goldstein from University of Michigan Department of Neurology. According to the ICPSR, ``The rationale for this study was to examine whether sleep, mood, and health related behaviors might differ between healthcare workers who transitioned to conducting care from home and those who continued to report in-person to their respective hospitals or healthcare facilities." The original data contained 916 survey responses. The average survey response answered 94.5\% of the survey questions, totalling 29 questions in total, with many questions leaving room for respondents to write how their mood or habits had changed since COVID-19 protocols were in effect. The data was stripped of any identifying information about the respondents and contained both categorical and numeric columns \cite{Web}. 

In this study, we would prefer not to use machine learning techniques, e.g. 
imputation, to treat the missing values for the categorical variables to keep the results more reliable. We directly remove the datapoints (rows) with many missing values from the dataset, and as the result, we end up with 518 data points. Since the mental health dataset contains categorical and ordinal variables, we first encode them to numbers. If it is necessary, we use several encoding techniques such as one-hot encoding or dummy variable encoding as well as several packages in Python such as \textit{OneHotEncoder}, \textit{LabelEncoder} or \textit{OrdinalEncoder} from \textit{sklearn.preprocessing} to prepare the dataset for analysis. 

\begin{figure}[H]
  \centering
    \caption{Q29 (a): ``Please tell us how your mood has changed.  My mood has been:''}    
  \includegraphics[width=.6\linewidth]{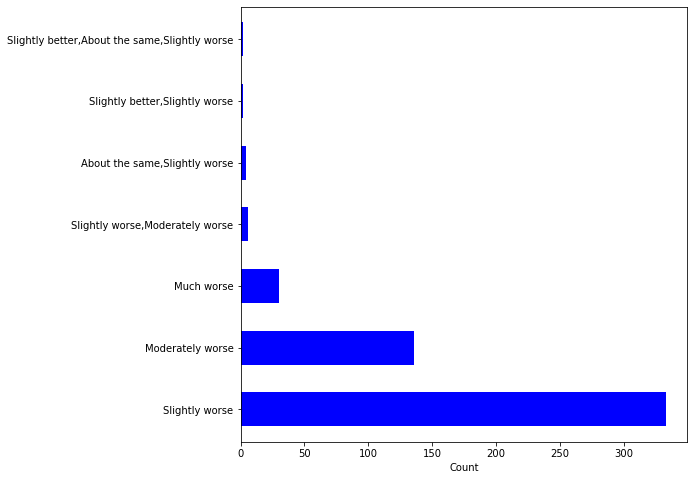}
  \label{fig:sub2212}
\end{figure}

In this paper, Question 29 (a) in the survey, which reads “Please tell us how your mood has changed?” (see Figure \ref{fig:sub2212}) is selected as the target variable. Our goal is to identify the top predictors of mental health decline by examining multiple machine learning models utilizing their feature importance. 
 
The remainder of this paper is structured as follows: Section 2 discusses the methodology, describes the experimental
framework used to find the top predictors of mental health decline among frontline workers in the United States. Section 3 discusses and analyzes the top predictors of mental health decline obtained by Machine Learning methods. Finally, Section 4 concludes the paper by summarizing our overall findings and suggests a direction for future research.

\section{Methods}
The data utilized in this study was taken from the University of Michigan’s Inter-University
Consortium for Political and Social Research, and collected by Deirdre Conroy \cite{ConroyDeirdre}. Conroy et al. \cite{conroy2021effects} confirms that all experiments were performed in accordance with relevant guidelines and regulations (see the Methods section in \cite{conroy2021effects}). The data is from a survey conducted within the University of Michigan Medical Center \cite{conroy2021effects}. All survey data was collected in accordance with relevant guidelines and regulations. The survey was undertaken after clearing University of Michigan Institutional Review Board (HUM00180147) approval, at which point the Qualtrics survey link was sent via email listservs that would reach large numbers of health care providers \cite{conroy2021effects}. It is noted that no compensation for participation was provided \cite{conroy2021effects}. Additionally, all participants have provided full and informed consent by completing the survey \cite{conroy2021effects}. 

\textbf{Data availability:} The data that support the findings of this study are available from the University of Michigan’s Inter-University Consortium for Political and Social Research, which is collected by Deirdre Conroy \cite{ConroyDeirdre,conroy2021effects}, at \url{https://www.openicpsr.org/openicpsr/project/127081/version/V1/view?path=/openicpsr/127081/fcr:versions/V1&type=project} \cite{Web}.
\subsection{Computational Process} 

Our main goal in this paper is to use multiple supervised and unsupervised machine learning models and techniques to find the top predictors (features) of mental health decline. Only for supervised machine learning models with high accuracy (at least 90\%), which are more reliable, we calculate the feature importance scores and proceed feature selection phase to find the top predictors for the mental health decline among frontline workers. Feature importance can be interpreted as the features that were the most valuable for generating the final prediction. Phrased differently, a high feature importance means the feature contains a lot of predictive power. Feature

In this section, we provide an overview and results for all supervised and unsupervised machine learning methods that we use in this paper. To recall, the goal of supervised learning is to predict accurately the value or the class of an unseen data point. To achieve this goal, we train a model on a training data and evaluate its accuracy on a test data. In all supervised methods, we split the data set into two parts, one part (for instance 75\% of all observations) as the training data set, and the other part (for instance 25\% of all data set) as the test data set. We train each model only on the training data set and then test the model on the test data set. Each supervised predictor contains some hyper-parameters that control overfitting or underfitting \cite{dietterich1995overfitting}. For instance, hyper-parameters for decision trees, k-nearest neighbors, neural networks are the depth of the tree, $k$, and the number of hidden layers, respectively. There are different  techniques, such as the elbow rule, for finding the best value for the hyper-parameters \cite{goodfellow2016machine}. Now, we overview the most common unsupervised and supervised techniques.

\subsubsection{Unsupervised Feature Selection}
\begin{figure}[h]
\centering
\begin{subfigure}{.48\textwidth}
  \centering
    \caption{Bar graph grouped by Question 29 (a) and Question 13}
  \includegraphics[width=\linewidth]{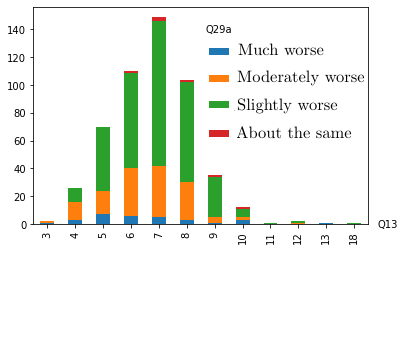}

  \label{fig:sub1}
\end{subfigure}%
\begin{subfigure}{.48\textwidth}
  \centering
    \caption{Bar graph grouped by Question 29 (a) and Question 21}
  \includegraphics[width=\linewidth]{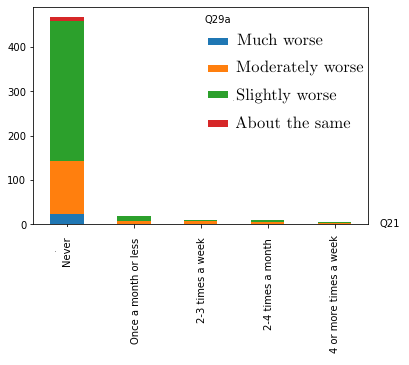}

  \label{fig:sub2}
\end{subfigure}
\caption{Bar graph grouped for the variables that pass the Chi-Squared test with significance level $\alpha=0.05$}
\label{fig:test1112}
\end{figure}

\begin{figure}[h]
\centering
\begin{subfigure}{.48\textwidth}
  \centering
    \caption{Bar graph grouped by Question 29 (a) and Question 22}
  \includegraphics[width=\linewidth]{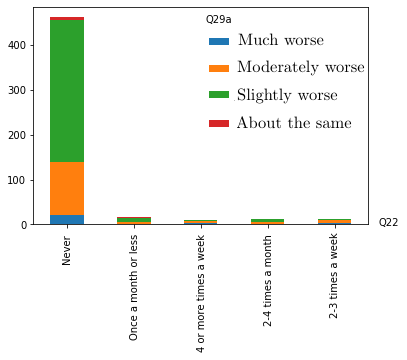}

  \label{fig:sub2}
\end{subfigure}
\begin{subfigure}{.48\textwidth}
  \centering
    \caption{Bar graph grouped by Question 29 (a) and Question 28}
  \includegraphics[width=\linewidth]{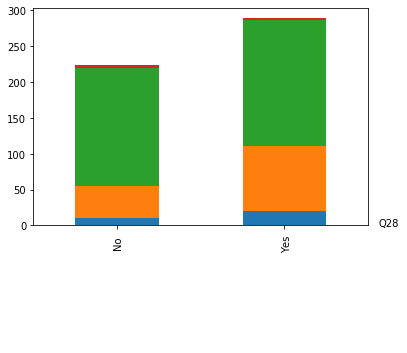}

  \label{fig:sub1}
\end{subfigure}%
\caption{Bar graph grouped for the variables that pass the Chi-Squared test with significance level $\alpha=0.05$}
\label{fig:test11}
\end{figure}

Since the problem is a classification problem, where the majority of variables are categorical, we can use statistical tests such as Chi-Squared tests to determine whether the target variable, Question 29 (a), is dependent or independent of the rest of variables. The variables that are independent can be considered as candidates for irrelevant features to the problem and they might be removed. During the data preparation process, we collapsed continuous variables into smaller groups (categories) to prepare the dataset for applying a Chi-Squared test \cite{mchugh2013chi}. After the data preparation process, we end up with a sample size of 513, which vastly exceeds the minimum of 20 to 50 recommended by Rana and Singhal \cite{rana2015chi} for avoiding Type II bias (failing to reject the null hypothesis when it is truly false). We first construct a contingency table, calculate the expected frequencies for pairs of target variables and each input variable, and then apply a Chi-Squared test with significance level $\alpha=0.05$ to determine whether or not there exists a correlation between frontline workers mental health decline and the rest of factors (features). It turns out that the Chi-Squared test rejects the null hypotheses, $H_0: \text{Question 29 (a)  is independent of Question $i$}$ versus the alternative $H_1: \text{Question 29 (a) is not independent of Question $i$}$, for $i \in \{13, 21, 22, 28\}$ (see Table \ref{chisqr}, and Figures \ref{fig:test1112} and \ref{fig:test11}). 
\begin{table}[H]
\scalebox{1}{%
\begin{tabular}[t]{ |l||l|  }
 \hline
 \multicolumn{2}{|c|}{Chi-Squared test rejects the null hypothesis that the target variable is independent of the follwoing variables} \\
\hline
\hline
 \multicolumn{2}{|l|}{Q13. Approximately how many hours did you sleep on an average work night in the last week?} \\
  \multicolumn{2}{|l|}{Q21. In January 2020, approximately how often did you use  marijuana/cannabis (recreational or medical)?} \\
    \multicolumn{2}{|l|}{Q22. In the last month, approximately how often did you use marijuana/cannabis (recreational or medical)?} \\
      \multicolumn{2}{|l|}{Q28. Has the amount of food you have been eating per day   changed?} \\
  
  \hline
\end{tabular}
}

\caption{Chi-Squared test with significance level $\alpha=0.05$ rejects the null hypotheses, $H_0: \text{Question 29 (a)  is independent of Question $i$}$ versus the alternative $H_1: \text{Question 29 (a) is not independent of Question $i$}$, for $i \in \{13, 21, 22, 28\}$}\label{chisqr}
\end{table}


\subsubsection{Supervised Feature Selection}
In this subsection, we provide an overview and results for several supervised machine learning methods.

 \textbf{Decision trees:} Decision trees are one of the most popular machine learning algorithms due to visualization simplicity. In other words, when we construct a tree, we can figure out what feature is most relevant at the first glimpse. A decision tree breaks down a dataset into smaller and smaller subsets containing data points that are more homogenous. Given a dataset with many features, different kind of decision trees with different depth can be constructed for training the model. However, finding the optimal decision tree, which is the shortest tree that can predict the class label for any unseen data, is computationally expensive. A decision tree contains a root node, internal nodes and leaf or terminal nodes. In a decision tree classifier, each leaf is assigned for a class label. Moreover, the root node and internal nodes contain feature test conditions. The order of non-terminal roots plays a significant role to find the optimal tree. A decision tree is also can be used for feature selection because it works based on the relevance of each feature. The goal of a decision tree is to minimize the entropy of the current sample to the next subsets by splitting the sample appropriately. The entropy for a sample $S$ with $n$ classes is defined as
$
E(S)=-\sum_{i=1}^{n}p_ilog({p_i}),
$
where $p_i$ is the probabilty of the class $i$ in the sample $S$\cite{kotsiantis2013decision}. A simple decision tree model is utilized on the Mental Health Data, and its accuracy turns out to be $74.12 \%$, which prevents us from analyzing the top predictors of the model. 

\textbf{Multinomial Logistic Regression:} Logistic regression is used to model binary categorical outcomes because it is improper to use linear regression while the output is not numeric, and the error terms are not normally distributed. In linear regression, parameters are estimated by ordinary least squares (OLS), where the sum of squared deviations of the predicted values from observed values is minimized. However, OLS is not appropriate for logistic regression to find unbiased estimators with minimum variance. For logistic regression, maximum likelihood estimation (MLE) is widely used to estimate parameters. Multinomial logistic regression is an appropriate extension of binary logistic regression for multiclass classification problems. It uses maximum likelihood estimation (MLE) to evaluate the probability of a target variable's classes \cite{bohning1992multinomial}. The accuracy of a multinomial logistic regression model on the Mental Health Data is $64.71 \%$, which is not satisfactory. 

\textbf{Recursive Feature Elimination:} The Recursive Feature Elimination (RFE) is a machine learning technique for finding the most relevant feature. It works by recursively removing attributes and building a model on those attributes that remain. If the accuracy of a model drops significantly in absence of a feature, it indicates that the feature is important. We apply the recursive feature elimination using multinomial logistic regression and decision tree in presence and absence of all features. Then we rank features according to the reduction of accuracy of a model in absence of a feature at a time.  Table \ref{RFE} displays the top predictors obtained by RFE using simple decision tree and multinomial logistic regression models.
 \begin{table}[H]
\scalebox{0.9}{%
\begin{tabular}[t]{ |c||c|  }
 \hline
 \multicolumn{2}{|c|}{RFE feature selection using Decision Tree} \\
 \hline
 Rank  &Feature\\
 \hline
  \#1&Q21:  In January 2020, approximately how often \\
 &  did you use marijuana/cannabis? \\
\hline
 \#2&Q24:  How many hours of COVID-19 related news    \\
 &or social media are you consuming on average per day? \\
 \hline
  \#3&Q20:  In the last month, approximately how often\\
   &did you have a drink containing alcohol? \\
 \hline
  \#4&Q23:  Has the amount of news you are consuming \\
 &  increased since the end of Feb, 2020?\\
 \hline
\end{tabular}
}
 \scalebox{0.9}{%
\begin{tabular}[t]{ |c||c|  }
 \hline
 \multicolumn{2}{|c|}{RFE feature selection using Multinomial Logistic Regression} \\
 \hline
 Rank  &Feature\\
  \hline
  \#1&Q18:  Has the amount of alcohol you  are consuming \\
 &  changed?\\
\hline
 \#2&Q14: Has the number of your worked hours\\
 &  per week changed? \\
 \hline
  \#3&Q16:   Have your sleep patterns changed?\\
&\\
 \hline
  \#4&Q20:  In the last month, approximately how often   \\
 &  did you have a drink containing alcohol?\\
 \hline
\end{tabular}
}
\caption{Applying RFE on a Decision Tree and a Multinomial Logistic Regression}
\label{RFE}
\end{table}

\textbf{Naive Bayes:} The Naive Bayes Classifier works based on the Bayesian theorem and is efficient when the dimensionality of the inputs is high. Given a data set comprising $N$ observations with their corresponding target class labels and $D$ features $X = (x_1,x_2,...,x_d)$, we want to find the posterior probability for the outcome of the event $\Omega$ among a set of possible outcomes $\{y_1,y_2,...,y_K$\}. The Naive Bayes classifier is assumed to be applied for datasets that the conditional probabilities of the features are statistically independent,
\begin{align}
p(\Omega = y_j|x_1,x_2,...,x_d) \propto p(x_1,x_2,...,x_d|y_j)p(y_j) 	\Longrightarrow & p(\Omega = y_j|x_1,x_2,...,x_d) \propto p(x_1|y_j)p(x_2|y_j)...p(x_d|y_j)p(y_j).
\end{align}

The Naive Bayes classifier reduces a high-dimensional task to a one-dimensional kernel density estimation. It is an effective and commonly used probabilistic machine learning classifier. Naive Bayes classifiers are especially used for text classification and spam detection \cite{domingos1997optimality}. There is very little training in Naive Bayes compared to other common classification methods. But, the accuracy of the Naive Bayes classifier model on the Mental Health Data is $43.52 \%$, which is significantly low.

 \textbf{k-Nearest Neighbors:} The k-nearest neighbors algorithm (KNN) classifies a new observation by a majority vote of its neighbors and it does not have the training phase. It calculates the distance from the new observation point to all seen data points, and then the new observation class label is assigned to be the class that is the most common among its $k$ nearest neighbors. KNN should be one of the first choices for a classification when there is little or no prior knowledge about the data and feature labels. {KNN} is computationally expensive, especially when the dimension of feature vector is high, because it must store all data points and their distances from the new observation \cite{jiang2007survey}. The accuracy of the KKN model on the Mental Health Data is $71.76 \%$.

 \textbf{Support Vector Machines:} The support vector machine algorithm seeks a hyperplane in an $n$-dimensional space ($n$ is the number of features) that classifies the data points with the maximum margin. The data points that are closer to the hyperplane and play an important role in finding the best position and orientation of the hyperplane to maximize the margine are called vector machines. One-vs-one Support Vector Machines (SVM OVO) is an appropriate extension of binary for multiclass classification problems. It splits the dataset into one dataset for each class versus every other class, which means it converts a multiclass classification dataset into multiple binary classification problems \cite{brereton2010support}. The accuracy of the SVM OVO model on the Mental Health Data is $70 \%$.

 \textbf{Neural Networks:} Neural networks are a set of algorithms that are designed to recognize patterns. We can consider them as clustering and classification layers on top of the data we store. They predict the label of unseen data according to similarities among the example inputs. Neural networks can also be used for extracting features that are fed to other algorithms for classification. Deep learning is the name that is used for stacked neural networks, which contains several layers. Each layer in the network is made of different neurons, where computations happen to decide whether a neuron should fire and the signal should progress further through the network. A neuron fires when it meets a sufficient stimuli. A neuron combines the last layer neurons output with a set of weights. This weighted sum is the input of a function called activation function to determine whether and to what extent the signal should progress to affect the ultimate outcome \cite{bishop2006pattern}. The accuracy of a neural network with 10 hidden layers and softmax as the last layer activation function on the Mental Health Data is $82.35 \%$. Hyperparameter tuning has also been done, but no better accuracy obtained. 

\begin{figure}[h]
  \centering
  \includegraphics[width=.65\linewidth]{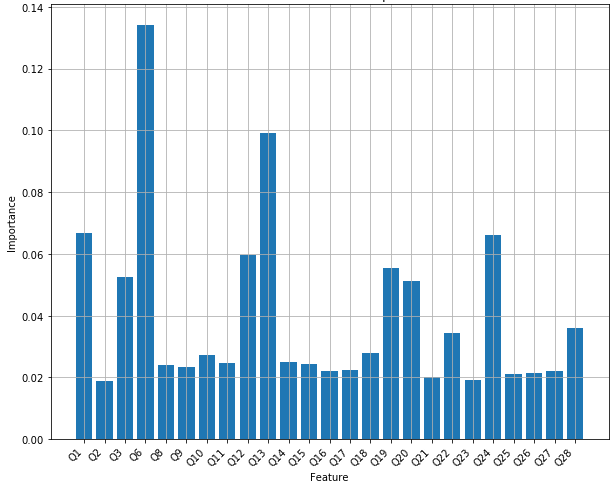}
  \caption{Feature importance scores of Random Forest with 10 trees with maximum depth of 10}
  \label{RandForestFeature}
\end{figure}

 \textbf{Random Forest:} Random Forest Classifier is an ensemble algorithm, which is a model that combines more than one algorithm of same or different kind for classifying objects. From randomly selected subset of training data, random forest generates a set of decision trees, and then it specifies the class label of a data point by aggregating the votes from different decision trees. The random forest is more powerful than a single decision tree classifier because it avoids overfitting on the training data. The accuracy of a regular random forest with 10 trees with maximum depth of 10 on the Mental Health Data turns out to be $80.59 \%$. Figure \ref{RandForestFeature} displays feature scores of the random forest model with 10 trees.

\begin{figure}[h]
  \centering
  \includegraphics[width=.7\linewidth]{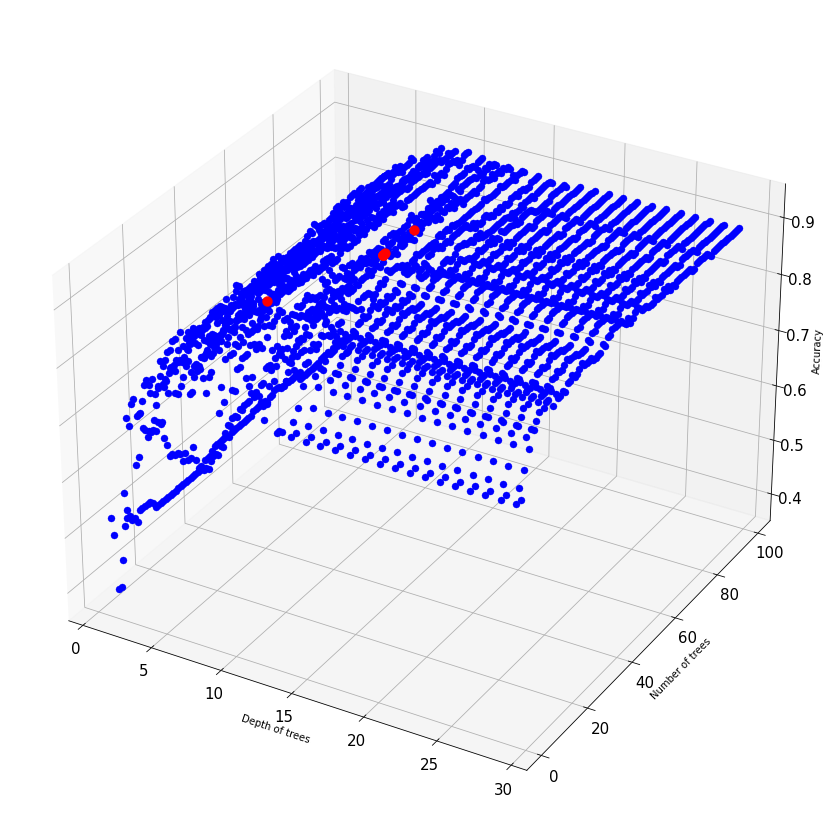}
  \caption{The accuracy scores of random forests with multiple values for its hyper-parameters. Red dots display the maximum accuracy of 92\% for random forest models using 15 trees with the maximum depth of 9, or 44 trees with the maximum depth of 12, or 45 trees with the maximum depth of 12, or 57 trees with the maximum depth of 12, or 60 trees with the maximum depth of 12, or 61 trees with the maximum depth of 12, or 63 trees with the maximum depth of 12. }
  \label{RandForesttune}
\end{figure}

\begin{figure}[h]
  \centering
  \includegraphics[width=.7\linewidth]{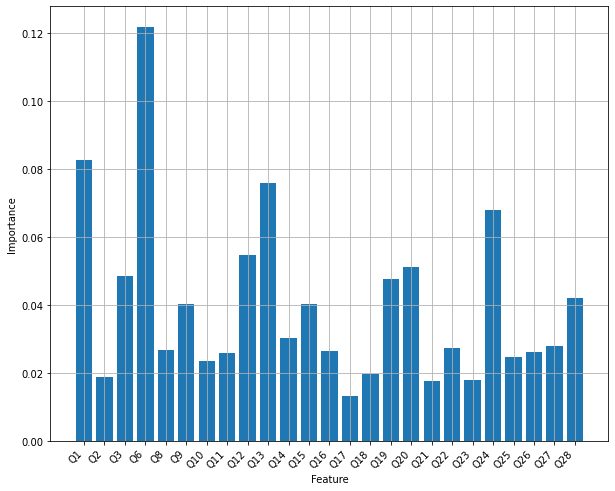}
  \caption{Feature importance scores of Random Forest with 44 trees with a maximum depth of 12}
  \label{RandForest4412}
\end{figure}

However, random forest hyperparameters, the number of trees and the maximum depth of each tree, play a very important role. To achieve the best accuracy of the model on the dataset, we need to tune hyper-parameters. We tune the number of trees between 1 and 100 while the maximum depth trees change between 1 and 30. It turns out that the maximum accuracy of a random forest is 92\% and it is obtained when it contains 15 trees with the maximum depth of 9, or 44 trees with the maximum depth of 12, or 45 trees with the maximum depth of 12, or 57 trees with the maximum depth of 12, or 60 trees with the maximum depth of 12, or 61 trees with the maximum depth of 12, or 63 trees with the maximum depth of 12. Figure \ref{RandForesttune} displays the accuracy scores of random forests with multiple values for its hyper-parameters. Figure \ref{RandForest4412} displays feature scores of the random forest model that utilizes 44 trees with a maximum depth of 12.

\begin{figure}[H]
\scalebox{.65}{
 \centering
  \includegraphics[width=\linewidth]{ 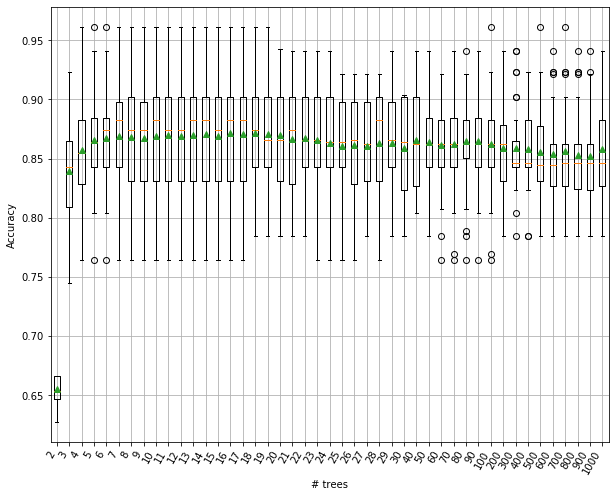}
  \caption{Box plot for the mean accuracy of Gradient Tree Boosting with different number of trees}
  \label{GradientTreeBoosting}
}
  \end{figure}

\begin{figure}[H]
\scalebox{.6}{
 \centering
  \includegraphics[width=.9\linewidth]{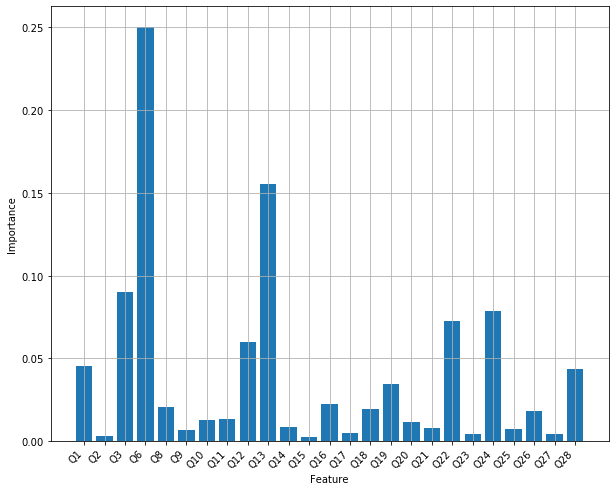}
  \caption{The feature scores of Gradient Tree Boosting with 19 trees}
  \label{GradientBoosting19trees}
  }
  \end{figure}

 \textbf{Gradient Tree Boosting:} Gradient tree boosting is an ensemble model, which is constructed from many individual decision tree models. Decision trees are added one at a time to the model and fit to minimize the error made by previous trees. In this model, gradient descent or stochastic gradient descent methods are used to minimize the differentiable loss functions. Gradient boosting is one of the most widely used machine learning algorithm on tabular datasets. The number of decision trees is one of the most important hyper-parameter for the Gradient Boosting ensemble algorithm. The depth of the trees and the number of trees can used efficiently in the ensemble to improve the performance of a model. Figure \ref{GradientTreeBoosting} displays a box and whisker plot and the effect of the number of trees on accuracy. It turns out that if we change the number of trees between 16 and 19, the mean accuracy scores of ensemble models do not change significantly. The mean accuracy of the ensemble model meets its maximum $87.1$\% when the number of trees is between 16 and 19. Figure \ref{GradientBoosting19trees} displays the feature scores Gradient Tree Boosting with 19 trees.
 
Gradient boosted trees have become the most widely used machine learning algorithms when it comes to tabular data. There are several popular boosting algorithms such as XGBoost, CatBoost and LightGBM. However, XGBoost, CatBoost and LightGBM algorithms differ from one another in the implementation of the boosted trees algorithm and the splitting mechanism. 

\begin{figure}[H]
\scalebox{1}{
 \centering
  \includegraphics[width=.9\linewidth]{ 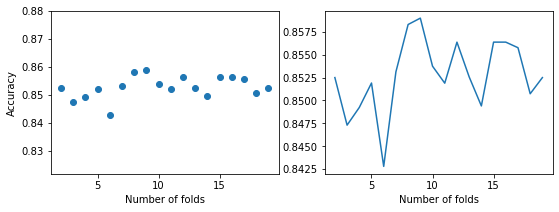}
  \caption{The mean accuracy of XGBoost using k-fold cross-validation}
  \label{XGBoost}
  }
  \end{figure}

\begin{figure}[H]
\scalebox{1}{
 \centering
  \includegraphics[width=.6\linewidth]{ 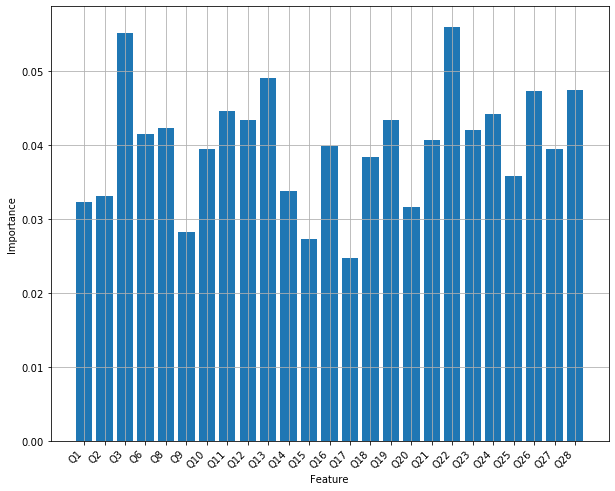}
  \caption{The feature scores of XGBoost using $k=9$ cross-validation}
  \label{XGBoost9folds}
  }
  \end{figure}

 \textbf{eXtreme Gradient Boosting (XGBoost):} XGBoost, introduced by Tianqi Chen \cite{chen2016xgboost,chen2015xgboost}, is a widely used machine learning algorithm that implements machine learning algorithms using the Gradient Boosting framework with high accuracy and speed. It was initially introduced to improve GBM’s training time. It provides a parallel tree boosting framework to solve so many multiclass classification problems.  Cross-Validation is a popular method to find a better accuracy for a model rather than a simple train/test split. The cross validation procedure is: shuffling the data set randomly, splitting the data set in $k$ groups, taking each group as a test data and feeding the rest to learn a model and find its accuracy, and then calculating and returning the mean of all $k$ obtained accuracy. If the value for $k$ is assigned to be the number of observations, then it is called leave-one-out cross-validation. Figure \ref{XGBoost} displays the accuracy scores of XGBoost with multiple folds ($2\leq k\leq 20$) for cross-validation. It turns out that XGBoost with k=9 folds gives the best accuracy, approximately 86\%. Figure \ref{XGBoost9folds} displays the feature scores of the XGBoost model using $k=9$ cross-validation.
  
\begin{figure}[H]
\scalebox{1}{
 \centering
  \includegraphics[width=.9\linewidth]{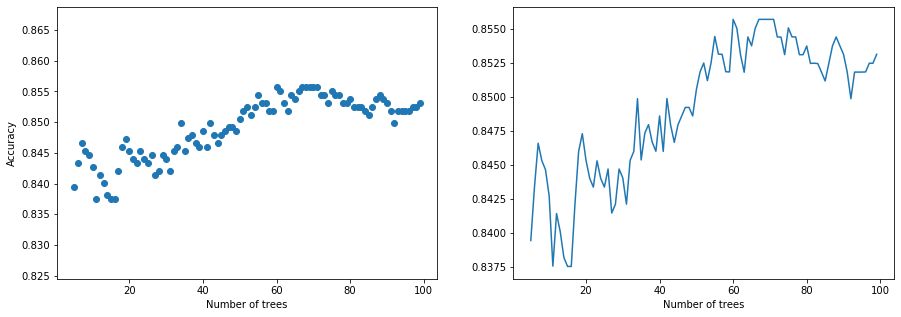}
  \caption{The mean accuracy of CatBoost using several trees}
  \label{CatBoost5-100}
  }
  \end{figure}

\begin{figure}[H]
\scalebox{1}{
 \centering
  \includegraphics[width=.6\linewidth]{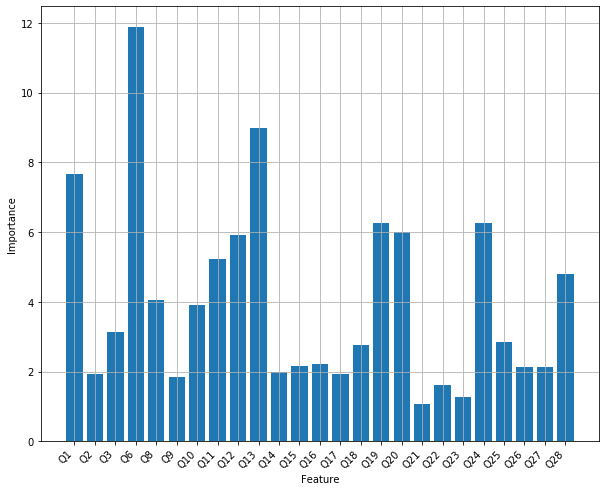}
  \caption{The feature scores of the CatBoost model using 60 trees}
  \label{CatBoost60trees}
  }
  \end{figure}

\textbf{CatBoost:} CatBoost \cite{dorogush2018catboost} is another algorithm for gradient boosting on decision trees. CatBoost trains trees sequentially, so that each successive tree is built with reduced loss compared to the previous trees. CatBoost can be used directly on categorical variables without any explicit pre-processing to convert categories into numbers. The number of trees is one of the most important hyper-parameters in CatBoost algorithm. Figure \ref{CatBoost5-100} displays the accuracy of CatBoost model on the Mental Health dataset for multiple number of trees. It turns out that CatBoost with 60 trees gives the best accuracy, approximately 85.50\%. Figure \ref{CatBoost60trees} displays the feature scores of the CatBoost model using 60 trees.

\textbf{Light Gradient Boosted Machine (LightGBM):} LightGBM, introduced by Guolin Ke et. al. \cite{ke2017lightgbm} in collaboration with Microsoft, is another strong learner that applies the boosting framework, which train trees sequentially instead of simultaneously, like CatBoost. LightGBM uses gradient-based one-side sampling (GOSS) that aims to find a good balance between increasing speed by reducing the number of data points and holding the accuracy for learned trees. LightGBM differs from XGBoost and CatBoost in the way that it works based on gradient-based one-side sampling (GOSS) and Exclusive Feature Bundling (EFB). 
 
GOSS excludes a significant proportion of data instances with small gradients, using the rest to estimate the information gain in individual trees. Guolin Ke et. al. \cite{ke2017lightgbm} prove the data instances with larger gradients play a more important role in the computation of information gain, and thus GOSS can obtain an accurate estimation of the information gain with a much smaller data size. EFB bundles mutually exclusive features to reduce the number of features.  Mutually exclusive features are features that rarely take nonzero values simultaneously, such as one-hot encoded features.  Guolin Ke et. al. \cite{ke2017lightgbm} prove that finding the optimal bundling of exclusive features is NP-hard, but a greedy algorithm can achieve a good approximation ratio, effectively reducing the number of features without hurting accuracy of the model.

\begin{figure}[H]
\centering
\begin{subfigure}{.47\textwidth}
  \centering
  \includegraphics[width=\linewidth]{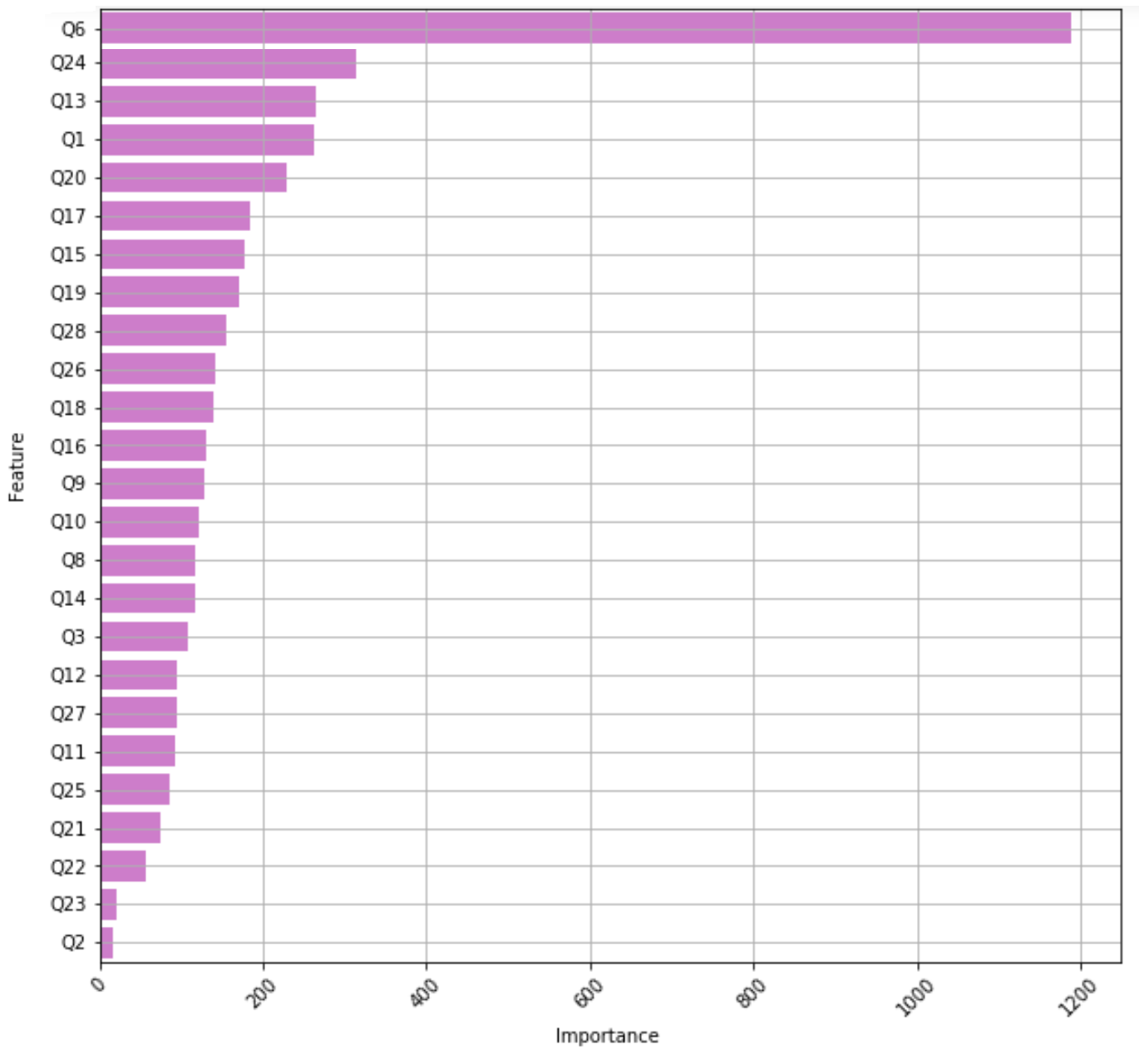}
  \caption{LightGBM stacked x1 variable importance}
  \label{fig:p2}
\end{subfigure}
\begin{subfigure}{.47\textwidth}
  \centering
  \includegraphics[width=\linewidth]{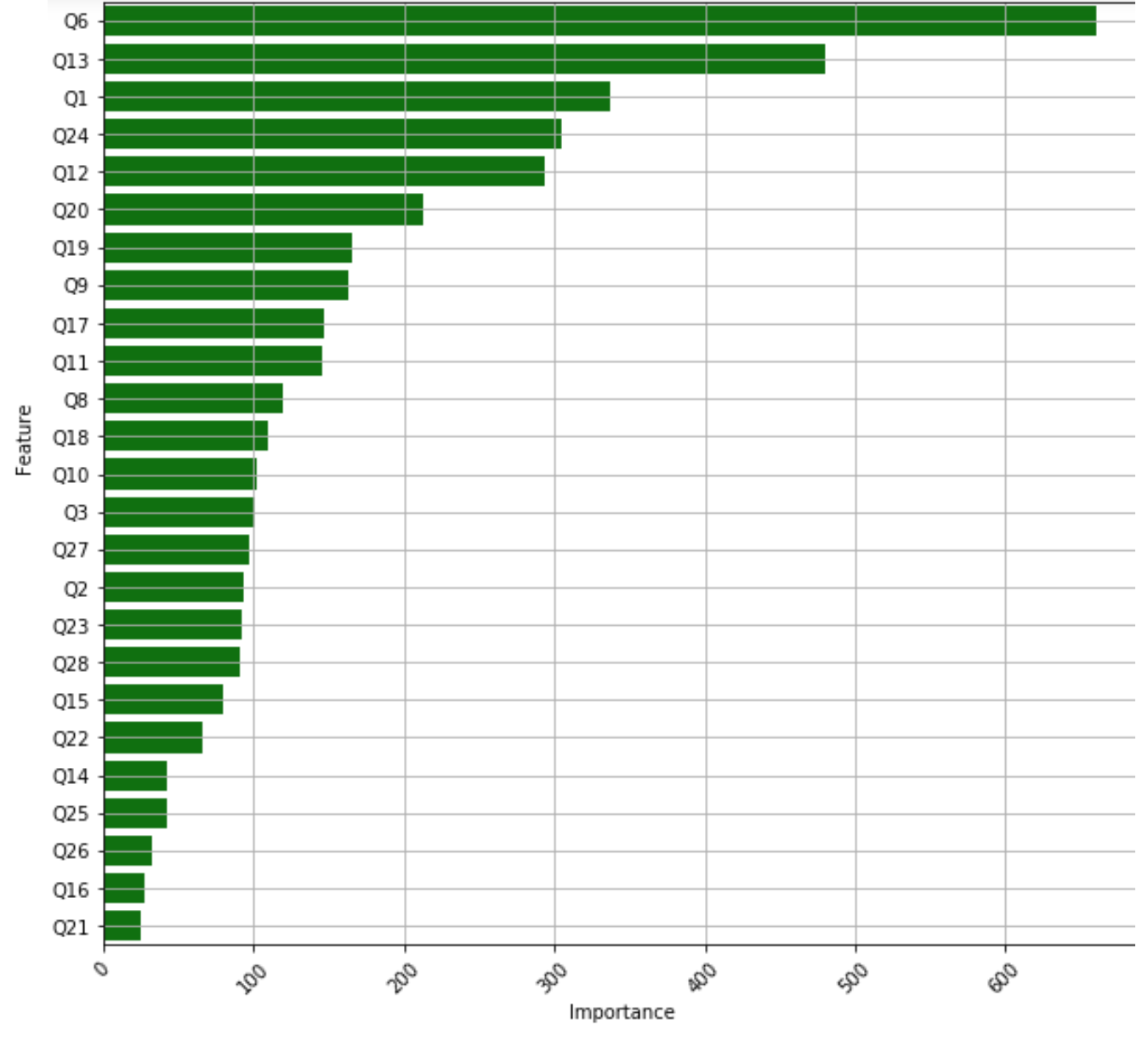}
  \caption{LightGBM stacked x2 variable importance}
  \label{fig:p2}
\end{subfigure}
\caption{LightGBM}
\label{fig:test123}
\end{figure}

Tables \ref{table2} and \ref{table3} display the accuracy scores of LightGBM model stacked once and twice, respectively. The model uses 200 gradient-boosted decision trees (GBDT) all limited to a maximum depth of 8. Figure \ref{fig:test123} displays the feature scores of LightGBM model stacked with one and two variables on the Mental Health data. The first four important features for LightGBM stacked with one and two variables are shown in Table \ref{lb1}.

\begin{figure}[H]
\begin{floatrow}
\capbtabbox{%
\scalebox{0.8}{%
  \begin{tabular}{ccccc } \hline
 & precision&recall&f1-score&support\\
 \hline
 accuracy &   & &0.81&154\\
 macro avg& 0.46  & 0.44   &0.43&154\\
 weighted avg& 0.77& 0.81&0.78&154\\
  \hline
  \end{tabular}
}
}
{%
\caption{ Stacked LightGBM Scores} \label{table2}
}
\capbtabbox{%
\scalebox{0.8}{%
  \begin{tabular}{ccccc } \hline
 & precision&recall&f1-score&support\\
 \hline
 accuracy &   & &0.91&180\\
 macro avg& 0.46 & 0.46   &0.46&180\\
 weighted avg& 0.90 & 0.91&0.90&180\\
  \hline
  \end{tabular}
}}{%
\caption{ 2x Stacked LightGBM Scores} \label{table3}
}
\end{floatrow}
\end{figure}

\begin{table}[H]
 \scalebox{.8}{%
\begin{tabular}[t]{ |c||p{83mm}|  }
 \hline
  \multicolumn{2}{|c|}{LightGBM stacked with 1 variables} \\
 \hline
 Rank  &Feature\\
  \hline
  \#1& Q6: What is your role in the healthcare field? (e.g. psychologist, physician, nurse)\\
\hline
 \#2&Q13:  Approximately how many hours did you sleep on an
average work night in the last week?\\
 \hline
  \#3&Q20: In the last month, approximately how often did you have a drink containing alcohol?\\
 \hline
  \#4&Q24: How many hours of COVID-19 related news or
social media are you consuming on average per day?\\
 \hline
\end{tabular}
}
\scalebox{.8}{%
\begin{tabular}[t]{ |c||p{83mm}|  }
 \hline
 \multicolumn{2}{|c|}{LightGBM stacked with 2 variables} \\
 \hline
 Rank  &Feature\\
  \hline
  \#1& Q6: What is your role in the healthcare field? (e.g. psychologist, physician, nurse)\\
\hline
 \#2&Q13: Approximately how many hours did you sleep on an average work night in the last week?\\
 \hline
  \#3&Q1: What is your age?\\
   &\\
 \hline
  \#4&Q24: How many hours of COVID-19 related news or
social media are you consuming on average per day?\\
 \hline
\end{tabular}
}
\caption{The most important variables for LightGBM}\label{lb1}
\end{table}

\begin{figure}[H]
\centering
\begin{subfigure}{.5\textwidth}
  \centering
  \includegraphics[width=.9\linewidth]{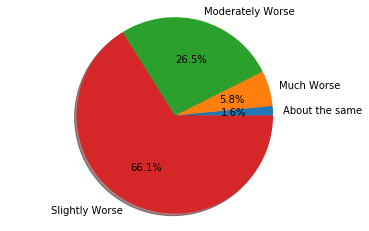}
  \caption{Before applying SMOTE, the distribution of examples across class labels of Question 29 (a) is not equal}
  \label{q29aaaabeforeSMOTE}
\end{subfigure}%
\begin{subfigure}{.5\textwidth}
  \centering
  \includegraphics[width=.85\linewidth]{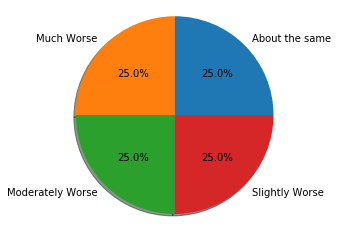}
  \caption{After applying SMOTE, the distribution of examples across class labels of Question 29 (a) is equal}
  \label{Piq251after}
\end{subfigure}
  \caption{Applying SMOTE to oversample all classes to the number of examples in the majority class}
\end{figure}

\begin{figure}[H]
  \centering
  \includegraphics[width=.6\linewidth]{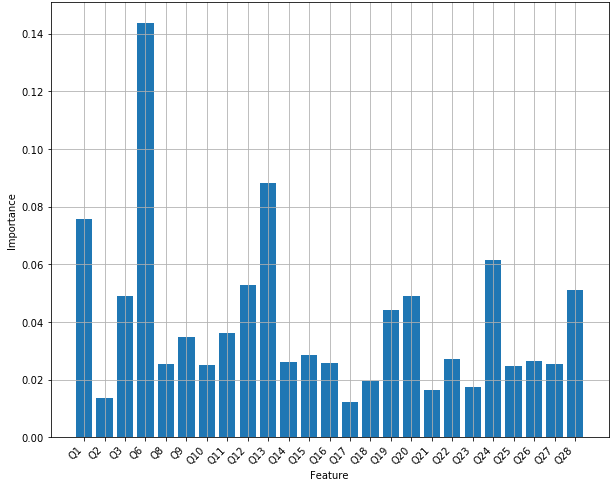}
  \caption{Feature importance scores of SMOTE Random Forest }
  \label{SMOTERAND}
\end{figure}

\begin{figure}[h]
  \centering
  \includegraphics[width=.8\linewidth]{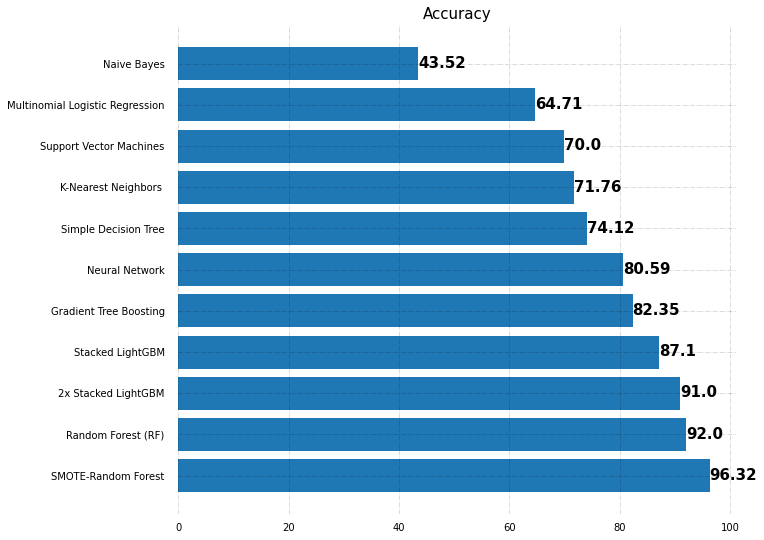}
  \caption{Improving the Randome Forest accuracy by means of SMOTE }
  \label{SMOTEcomparison}
\end{figure}

\textbf{Synthetic Minority Oversampling Technique:} Note that the categories of the target variable, Question 29 (a), are not approximately equally represented. Nitesh Chawla, et al. \cite{chawla2002smote} proposed a technique called the Synthetic Minority Oversampling Technique, or SMOTE, for synthesizing new examples of the minority classes. The SMOTE is a combination of  over-sampling the minority classes and under-sampling the majority classes to achieve a better classifier performance. Since there are few examples with labels belong to some classes, we remove them before we apply SMOTE to avoid misleading the model (see Figure \ref{q29aaaabeforeSMOTE}). Applying the SMOTE (see Figure \ref{Piq251after}) to oversample all classes to the number of examples in the majority class on Random Forest improves accuracy of the model to $96.32 \%$ (see Figure \ref{SMOTEcomparison}) with questions 6, 13, 1 and 24 as the most important features, respectively (see Figure \ref{SMOTERAND}). The first four important features for the SMOTE Random Forest are shown in Table \ref{SMOTEFEATURE}.

\begin{table}[h]
\begin{tabular}[t]{ |c||p{103mm}|  }
 \hline
  \multicolumn{2}{|c|}{SMOTE Random Forest} \\
 \hline
 Rank  &Feature\\
  \hline
  \#1& Q6: What is your role in the healthcare field? (e.g. psychologist, physician, nurse)\\
\hline
 \#2&Q13:  Approximately how many hours did you sleep on an
average work night in the last week?\\
 \hline
  \#3&Q1: What is your age?\\
 \hline
  \#4&Q24: How many hours of COVID-19 related news or
social media are you consuming on average per day?\\
 \hline
\end{tabular}
\caption{The first four most important features for the regular SMOTE Random Forest}\label{SMOTEFEATURE}
\end{table}

\begin{figure}
\begin{center}
\scalebox{1}{%
\tikzstyle{decision} = [diamond, draw, fill=blue!20, 
    text width=4.5em, text badly centered, node distance=3cm, inner sep=0pt]
\tikzstyle{block} = [rectangle, draw, fill=blue!20, node distance=4cm,
    text width=15em, text centered, rounded corners, minimum height=3em]
\tikzstyle{blockkk} = [rectangle, draw, fill=blue!20, node distance=6cm,
    text width=10em, text centered, rounded corners, minimum height=3em]
\tikzstyle{blockk} = [rectangle, draw, fill=green!10, node distance=9cm,
    text width=30em,  rounded corners, minimum height=10em]
\tikzstyle{line} = [draw, -latex']
\tikzstyle{cloud} = [draw, ellipse,fill=red!20, node distance=6cm,
    minimum height=3em]    
\begin{tikzpicture}[node distance = 3cm, auto]
    \node [block] (init) {The data from the University of Michigan's Inter-university Consortium for Political and Social Research (ICPSR); Select Question 29 (a) as the target variable};
    \node [block, below of=init] (identify) {Data Preprocessing such as dealing with missing values and encoding process};
    \node [cloud, right of=identify] (expert) {Unsupervised Learning};
    \node [cloud, left of=identify] (system) {Supervised Learning};
    \node [block, below of=system] (identify2) {Split the data into training and test sets};
    \node [blockkk, right of=identify2] (identify221) {Recursive Feature Elimination using a simple decision trees and a multinomial logistic regression};
    \node [block, below of=identify2] (identify21) {Apply several supervised machine learning models including a simple decision trees, multinomial logistic regression, naive bayes, k-nearest neighbors, support vector machines, neural networks, random forests, gradient tree boosting, XGBoost, CatBoost, LightGBM, and random forest using Synthetic Minority Oversampling};
    \node [block, below of=identify21] (identify211) {Figure \ref{SMOTEcomparison} displays the accuracy scores of all supervised machine learning models that have been utilized in this study};
    \node [block, below of=expert] (identify3) {Chi-Squared test with significance level $\alpha=0.05$ rejects the null hypotheses, $H_0$: Question 29 (a)  is independent of Question $i$ versus the alternative $H_1$: Question 29 (a) is not independent of Question $i$, for $i \in \{13, 21, 22, 28\}$};
 \node [blockk, right of=identify21] (Final) {\textbf{The most important questions ( the top predictors of mental health decline):} \\
Question 6. ``What is your role in the healthcare field?” \\
Question 13. ``Approximately how many hours did you sleep on an average work night in the last week?”\\
Question 24. “How many hours of COVID-19 related news or social media are you consuming on average per day?” \\
Question 1. “What is your age?”\\
Question 20. ``In the last month, approximately how often did you have a drink containing alcohol?"};

    \path [line] (init) -- (identify);
    \path [line] (identify) --(expert);
    \path [line] (identify) -- (system);
    \path [line] (expert) -- (identify3);
    \path [line] (system) -- (identify2);
    \path [line] (identify2) -- (identify221);
\path [line] (identify221) --(Final);
    \path [line] (identify2) -- (identify21);
    \path [line] (identify21) -- (identify211);
    \path [line] (identify211) -|(Final);
    \path [line] (identify3) -| (Final);

\end{tikzpicture}
}
\end{center}
\caption{Multiple statistical and machine learning models and techniques such as Decision Trees, Multinomial Logistic Regression, Naive Bayes, k-Nearest Neighbors, Support Vector Machines, Neural Networks, Random Forests, Gradient Tree Boosting, XGBoost, CatBoost, LightGBM, Synthetic Minority Oversampling, and a Chi Squared Test have been used to identify the most important factor in predicting the mental health decline of a frontline worker. It turns out that the top predictors are the healthcare role the individual is in (Nurse, Emergency Room Staff, Surgeon, etc.), followed by the amount of sleep the individual has had in the last week, the amount of COVID-19 related news an individual has consumed on average in a day, the age of the worker, and the usage of alcohol and cannabis.}\label{Flowchart}
\end{figure}
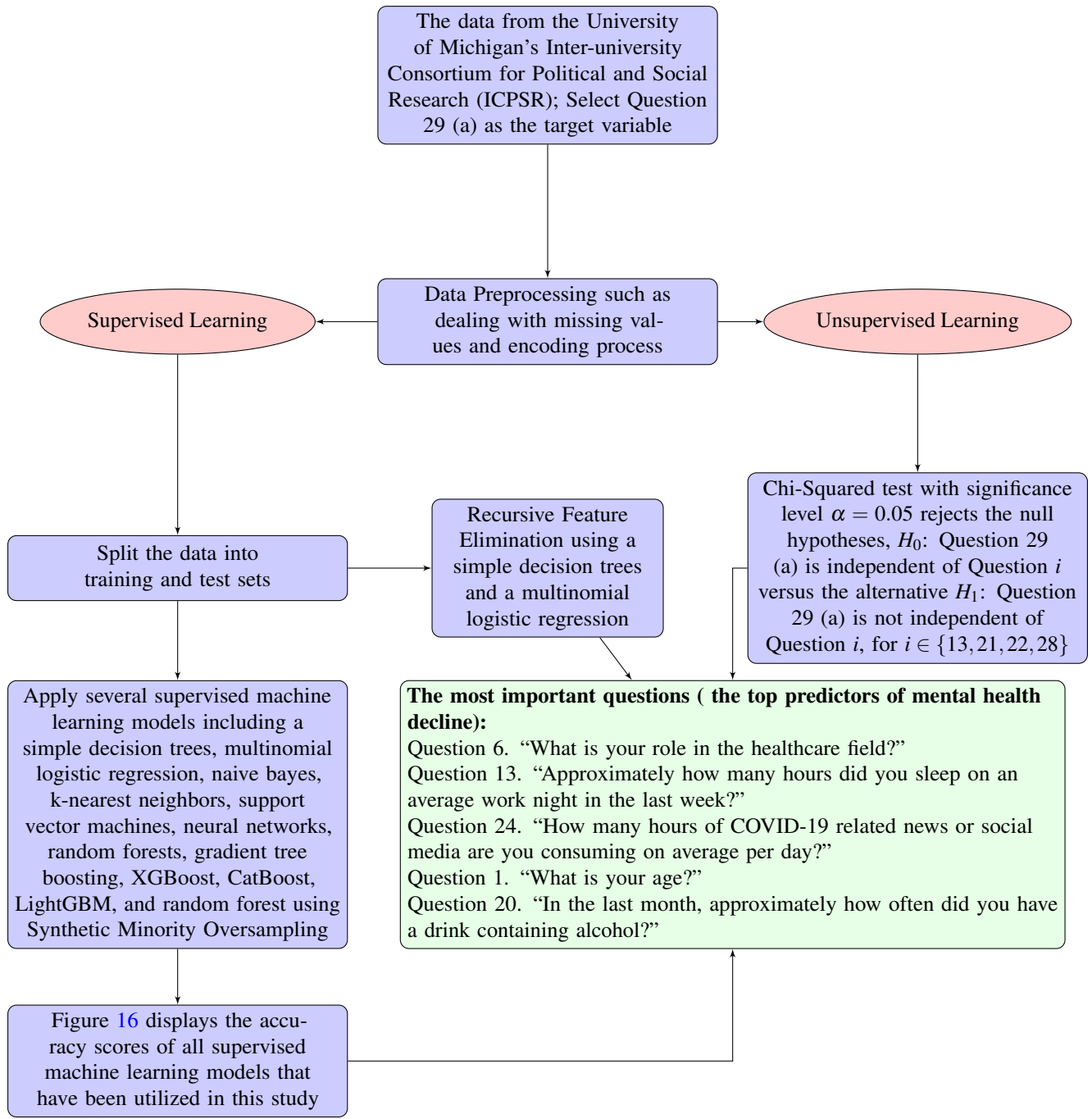

Figure \ref{SMOTEcomparison} displays the accuracy scores of all supervised machine learning models that discussed in this section. Top predictors for mental health analysis have been identified from different approaches. Among all the approaches, the random forest using SMOTE is the model that has identified the maximum top predictors. Figure \ref{Flowchart} summarizes the methology and the results in this section. 
\section{Discussion}
Upon the examination of the top predictors from the many different models, we have identified four main questions as well as a group of questions (questions 18-22) as highly predictive of mental health decline in frontline workers. With the emergence of new strains of COVID-19 such as the Delta and Omicron variants, we may be able to learn from our past and improve the working environment for frontline workers such that their mental health does not decline much, if at all.

Our analysis concludes that Question 6: ``What is your role in the healthcare field?” is the most important predictor of mental health decline. Both stacked LightGBM models, the Random Forest with and without SMOTE, CatBoost, and Gradient Tree Boosting all conclude that this is the most important predictor. This result is unsurprising, as we do not expect a frontline worker dealing with extremely sick COVID-19 patients would suffer a greater mental health decline than a child pediatrician who is not interacting with COVID-19 patients. One could find in the face of the recent variants and subvariants, such as the Omicron variant and its subvariants, that these issues may flare up again, and there may be no one size fits all solution for the medical institution seeking to preserve the mental health of its employees across different disciplines. As such, implementing role specific support groups may be able to relieve some of the stress different workers may be facing, as sharing experiences with someone who has gone through or is currently experiencing the same things may allow for the workers to bond over their adversities and preserve their mental health.

We also see a clear second most important predictor in Question 13: ``Approximately how many hours did you sleep on an average work night in the last week?” The Random Forest with and without SMOTE, Gradient Tree Boosting, CatBoost, both LightGBM models, and the Chi Squared analysis find this is a highly important predictor. We know that COVID-19 if contracted can have large, potentially deadly, effects on frontline workers. However, this analysis reveals one of the “hidden” effects COVID-19 has as well. Even without contracting the virus, frontline workers are having impacted sleeping schedules, as the stress of their jobs takes a toll on their body both mentally and physically. This may also be part of the institutions they are employed by asking for longer hours to compensate for a lack of frontline workers or an excess of patients sick with the virus. The institutions may find that the frontline workers' performance and level of care may increase if they resume their normal sleep schedules as well as the mental health of their workers improving. 

Our analysis finds that Question 24: “How many hours of COVID-19 related news or social media are you consuming on average per day?” is a highly important predictor of mental health decline in frontline workers. The Random Forest with and without SMOTE, Gradient Tree Boosting, both LightGBM models, and the Recursive Feature Elimination utilizing a decision tree all find this is in the top 5 most important predictors. At first, we found this result surprising, but after further deliberation, we believe that this may be the easiest factor to improve on. There is seldom positive news regarding the virus circulating on the news or on social media, with many headlines pointing to new variants that are more contagious or deadly than the one before it, or how we are breaking new daily case records, and even that other states and countries are re-entering lockdowns. It is not surprising to see that consuming more and more of this media can lead one to doubt that we will ever beat COVID-19 and that the fight may be futile, leading to a decrease in mental health. One could purposefully try to limit their consumption of such media and may find that their mental health and possibly even physical health (such as sleep patterns) may improve.

Looking at the top predictors of the Random Forest with and without SMOTE, CatBoost, and both LightGBM models, we find that Question 1: “What is your age?” is one of the top five most important predictors of mental health decline in frontline workers. We hypothesize that workers that are younger and less familiar with intense medical situations may be more prone to experiencing mental health decline upon encountering said situations. Healthcare institutions may want to encourage their less and more seasoned staff to work together, allowing for the younger staff to learn helpful tips for their job as well as enabling them to reach out to the seasoned staff if they feel as though they are beginning to have mental health issues.

Lastly, we find that a group of questions regarding alcohol and cannabis use stand out as highly predictive in many different models such as the Recursive Feature Elimination with both Decision Trees and Multinomial Regression, the Chi Squared test, the Random Forest with and without SMOTE, XGBoost, CatBoost, and both LightGBM models. This result is intuitive, as many people may turn to different methods of coping with the extreme stress that they may be feeling during their day-to-day lives combating COVID-19 and its many variants. Healthcare institutions may find that offering counseling services and other stress relief programs may decrease the usage of Alcohol and Cannabis in their frontline workers, improving both their mental and physical health. 

Our analysis has demonstrated that COVID-19 has not taken a toll on the mental health of frontline workers, but their physical health too. With the emergence of many variants and subvariants, large healthcare institutions can learn from the past to improve conditions and offer their frontline workers a better chance of improving/maintaining their mental and physical health. We know some of the physical symptoms that come from contracting COVID-19, but what we have found is that there exist some “hidden” effects of the virus that do not necessarily come from direct contact. These “hidden” effects may be partially preventable, and in the face of new variants that could cause a resurgence of the virus, healthcare institutions must take steps to prevent these “hidden” symptoms from becoming problems for our frontline workers. Some possible remedies may be easier to implement than others, such as warning staff about consuming too much COVID-19 related news and offering counseling. Even working to allow younger staff to work with older, more experienced staff may result in better training for the institution while simultaneously working to better their staff’s combined mental health.
\section{Conclusion and Future Work}

We have presented an analysis of a COVID-19 mental health survey data obtained from the University of Michigan Inter-University Consortium for Political and Social Research. As part of the analysis, we have utilized a variety of statistical and machine learning models and techniques such as Decision Trees, Multinomial Logistic Regression, Naive Bayes, k-Nearest Neighbors, Support Vector Machines, Neural Networks, Random Forests, Gradient Tree Boosting, XGBoost, CatBoost, LightGBM, Synthetic Minority Oversampling, and a Chi Squared Test. Through the interpretation of the many models applied to the mental health survey data, we have concluded that the most important factor in predicting the mental health decline of a frontline worker is the healthcare role the individual is in (Nurse, Emergency Room Staff, Surgeon, etc.), followed by the amount of sleep the individual has had in the last week, the amount of COVID-19 related news an individual has consumed on average in a day, the age of the worker, and the usage of alcohol and cannabis. In light of the recent identification of the Omicron and Delta variants of COVID-19, we hope that these findings can be utilized by healthcare facilities to help preserve or improve their employee’s mental health.

In future work we would like to aggregate more data on frontline workers, their habits, and their mental health. With more data from a diverse range of locations, we would have the ability to apply even more complex and accurate models to the data, while simultaneously allowing us to make even stronger conclusions about the impacts COVID-19 has on the mental health of frontline workers. Additionally, in future work, we may be able to utilize permutation importance along with the feature importance to shed even more light on the inner workings of the matching learning models. Finally, we would like to identify specific features then utilize generalized linear models to more exactly quantify the relationship the variables have with mental health decline. 

We are also intesred in analyzing accuracy, speed and feature scores of some machine learning models on the COVID-19 mental health data when we replace the most common optimizers such as GD, SGD or Limited-memory BFGS with the ones that are introduced recently \cite{rezapour2019adaptive,erway2021new}.

\end{document}


\appendix
\section{Some important features in the mental health data}\label{Appendix}
\begin{table}[H]
\centering
\scalebox{.8}{
\begin{tabular}{|p{40mm}|p{35mm}||p{40mm}|p{35mm}|}
  \hline
 Question & Levels & Question & Levels  \\
  \hline
  Q1: What is your age? &  10-20\newline 21-30\newline 31-40\newline  41-50\newline 51-60\newline 61-70\newline  71-80\newline  81-90& Q6: What is your role in the healthcare field?  &  Psychologist \newline   Physician\newline Nurse\newline Psychiatrist\newline Social Worker \newline Research coordinator/manager/Public health/Staff\newline etc.\\
  \hline
    Q9:  Are you conducting clinical video visits with your patients from your home?&  No\newline Yes & Q10:  Does your work require you to follow fixed work schedule while working from home?&  No\newline Yes \\
  \hline
      Q14:   Has the number of your worked hours per week changed?&  No\newline Yes & Q15:  Have you varied your work schedule?&  No\newline Yes \\
  \hline
        Q16:   Have your sleep patterns changed?&  No\newline Yes& Q17:   Has the number of naps you are taking changed?&  No\newline Yes \\
  \hline
        Q18:  Has the amount of alcohol you are consuming changed?&  No\newline Yes&  Q19:  In January 2020, approximately how often did you have a drink containing alcohol?&  4 or more times a week\newline2-4 times a month\newline 2-3 times a week \newline Once a month or less \newline Never\\
  \hline
          Q20:  In the last month, approximately how often did you have a drink containing alcohol?&  4 or more times a week\newline2-4 times a month\newline 2-3 times a week \newline Once a month or less \newline Never& Q21:  In January 2020, approximately how often did you use marijuana/cannabis?&  4 or more times a week\newline2-4 times a month\newline 2-3 times a week \newline Once a month or less \newline Never\\
  \hline
\end{tabular}
}
\end{table}

\begin{table}[H]
\centering
\scalebox{.8}{
\begin{tabular}{|p{40mm}|p{35mm}||p{40mm}|p{35mm}|}
  \hline
 Question & Levels & Question & Levels \\
  \hline
Q22:  In the last month, approximately how often did you use marijuana/cannabis?&  4 or more times a week\newline2-4 times a month\newline 2-3 times a week \newline Once a month or less \newline Never& Q23:  Has the amount of news you are consuming increased since the end of Feb, 2020?&  No \newline Yes\\
  \hline
 Q24:  How many hours of COVID-19 related news or social media are you consuming on average per day? &  4 or more times a week\newline2-4 times a month\newline 2-3 times a week \newline Once a month or less \newline Never&Q25: Have you had more ``screen time” e.g. use of smartphone, tablet, etc. around bedtime?&  No \newline Yes\\
 \hline
 Q26: Have you changed your movement/exercise? &  No \newline Yes&Q27: Has the quality of your diet changed? &  No \newline Yes\\
 \hline
 Q28: Has the amount of food you have been eating per day changed? &  No \newline Yes&Q29: Has your mood changed? &  No \newline Yes\\
 \hline
  Q8:  Are you currently conducting your job mostly from home now?&  No\newline Yes &Q29 (a): Please tell us how your mood has changed.  My mood has been: &  Much better \newline  Moderately better\newline Slightly better \newline About the same\newline Slightly worse \newline Moderately worse\newline Much worse\\
 \hline
 
\end{tabular}
}
\end{table}

\begin{figure}
\centering
\begin{subfigure}{.33\textwidth}
  \centering
  \includegraphics[width=1\linewidth]{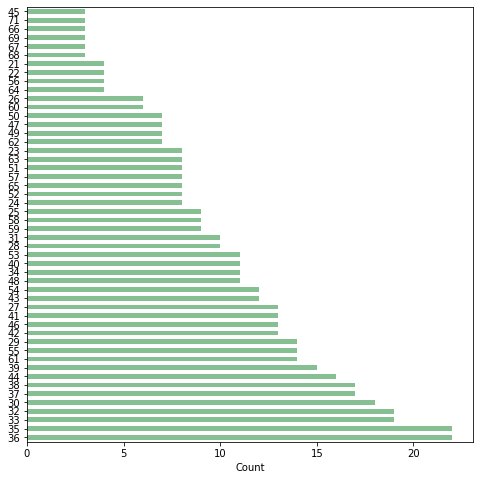}
  \caption{Q1: What is your age?}
  \label{fig:sub1}
\end{subfigure}%
\begin{subfigure}{.33\textwidth}
  \centering
  \includegraphics[width=1\linewidth]{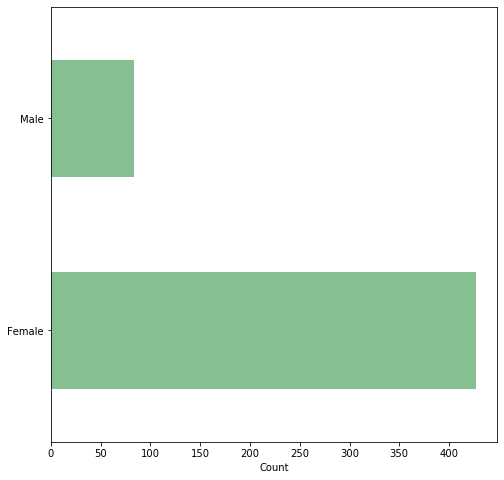}
  \caption{Q2: What is your gender?}
  \label{fig:sub2}
\end{subfigure}
\begin{subfigure}{.33\textwidth}
  \centering
  \includegraphics[width=1\linewidth]{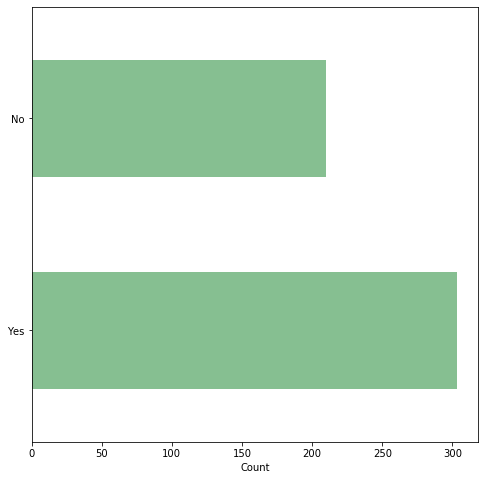}
  \caption{Q8:  Are you currently conducting your job mostly from home now?}
  \label{fig:sub2}
\end{subfigure}

\begin{subfigure}{.33\textwidth}
  \centering
  \includegraphics[width=1\linewidth]{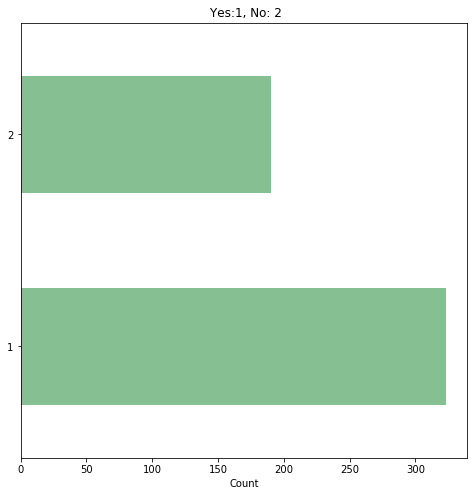}
  \caption{Q9:  Are you conducting clinical video visits with your patients from your home?}
  \label{fig:sub1}
\end{subfigure}%
\begin{subfigure}{.33\textwidth}
  \centering
  \includegraphics[width=1\linewidth]{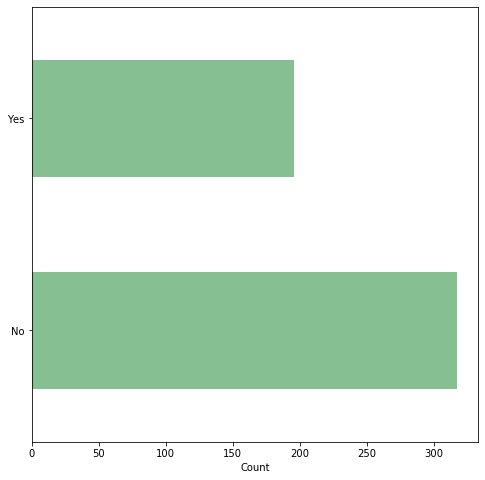}
  \caption{Q10:  Does your work require you to follow a fixed work schedule while working from home?}
  \label{fig:sub2}
\end{subfigure}
\begin{subfigure}{.33\textwidth}
  \centering
  \includegraphics[width=1\linewidth]{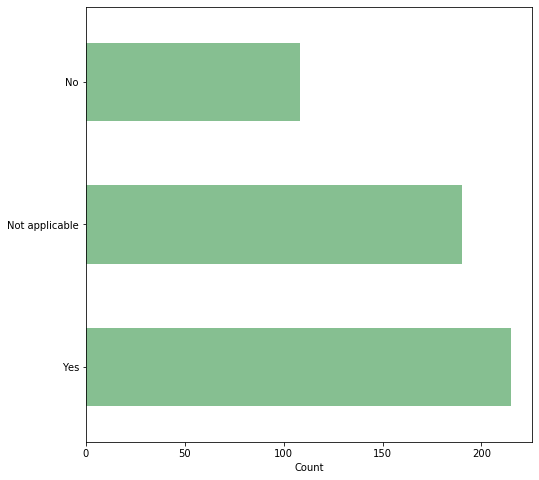}
  \caption{Q11:   Are children home from school in the house?}
  \label{fig:sub2}
\end{subfigure}

\begin{subfigure}{.33\textwidth}
  \centering
  \includegraphics[width=1\linewidth]{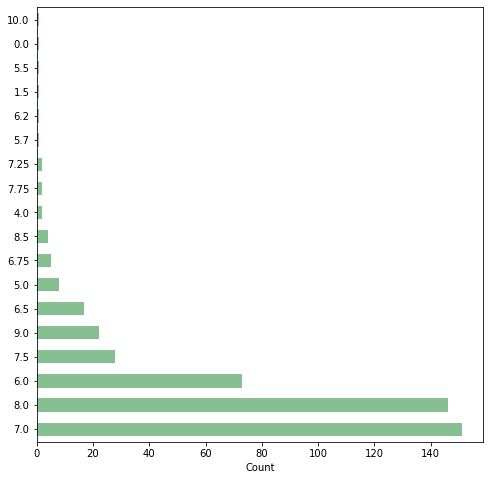}
  \caption{Q12:  Approximately how many hours did you sleep on an average work night in January 2020?}
  \label{fig:sub1}
\end{subfigure}%
\begin{subfigure}{.33\textwidth}
  \centering
  \includegraphics[width=1\linewidth]{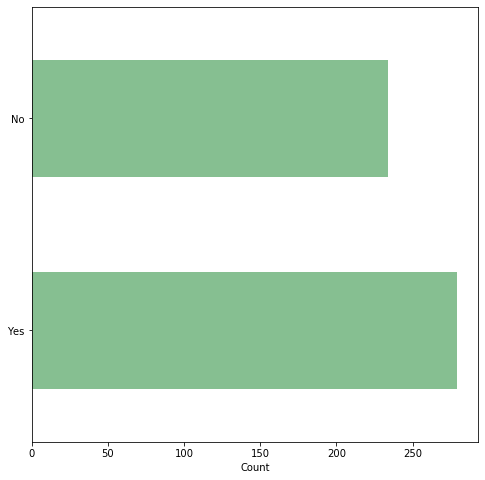}
  \caption{Q14:  Has the number of your work hours per week changed?}
  \label{fig:sub2}
\end{subfigure}
\begin{subfigure}{.33\textwidth}
  \centering
  \includegraphics[width=1\linewidth]{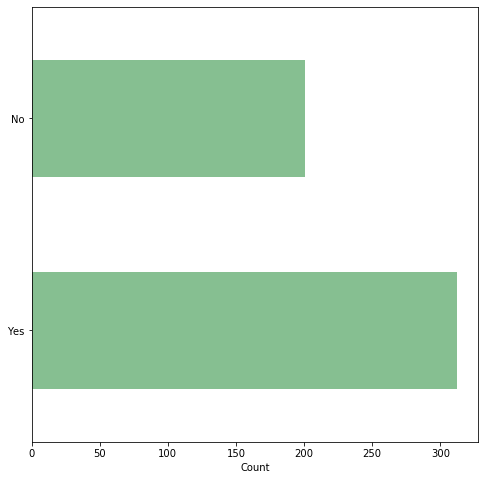}
  \caption{Q15:  Have you varied your work schedule?}
  \label{fig:sub2}
\end{subfigure}
\caption{Frequency bar charts after the preprocessing phase}
\label{fig:test}
\end{figure}

\begin{figure}
\centering
\begin{subfigure}{.33\textwidth}
  \centering
  \includegraphics[width=1\linewidth]{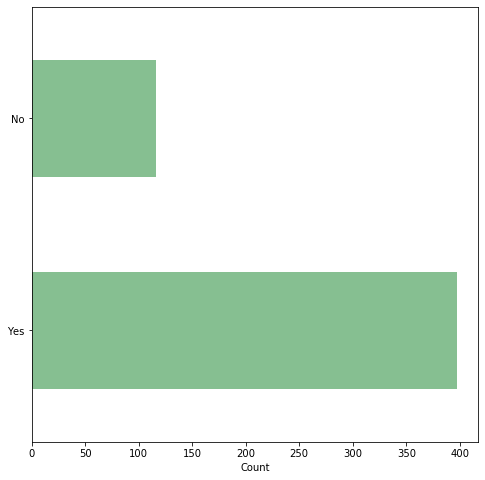}
  \caption{Q16:   Have your sleep patterns changed?}
  \label{fig:sub1}
\end{subfigure}%
\begin{subfigure}{.33\textwidth}
  \centering
  \includegraphics[width=1\linewidth]{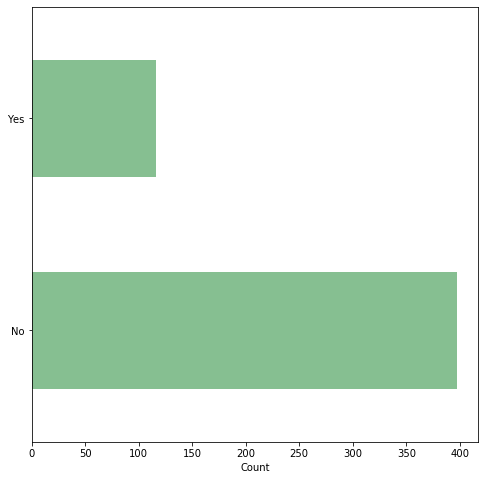}
  \caption{Q17:  Has the number of naps you are taking changed?}
  \label{fig:sub2}
\end{subfigure}
\begin{subfigure}{.33\textwidth}
  \centering
  \includegraphics[width=1\linewidth]{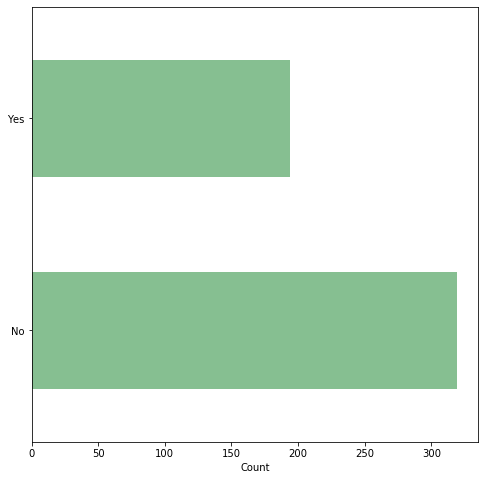}
  \caption{Q18:  Has the amount of alcohol you are consuming changed?}
  \label{fig:sub2}
\end{subfigure}

\begin{subfigure}{.33\textwidth}
  \centering
  \includegraphics[width=1\linewidth]{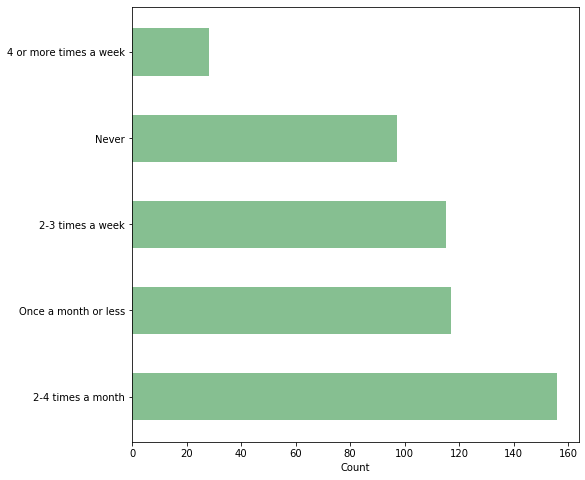}
  \caption{Q19:  In January 2020, approximately how often did you have a drink containing alcohol?}
  \label{fig:sub1}
\end{subfigure}%
\begin{subfigure}{.33\textwidth}
  \centering
  \includegraphics[width=1\linewidth]{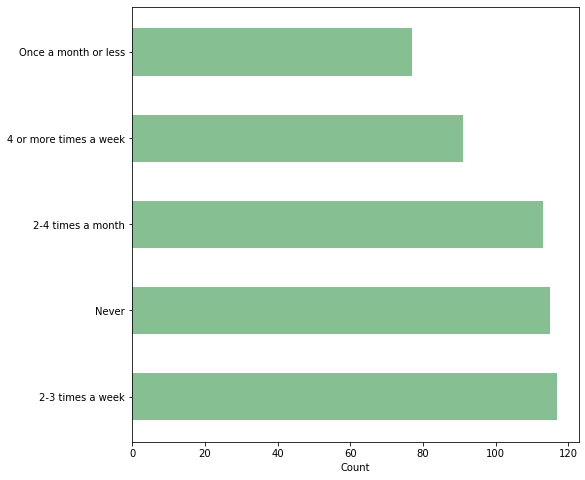}
  \caption{Q20:  In the last month, approximately how often did you have a drink containing alcohol?}
  \label{fig:sub2}
\end{subfigure}
\begin{subfigure}{.33\textwidth}
  \centering
  \includegraphics[width=1\linewidth]{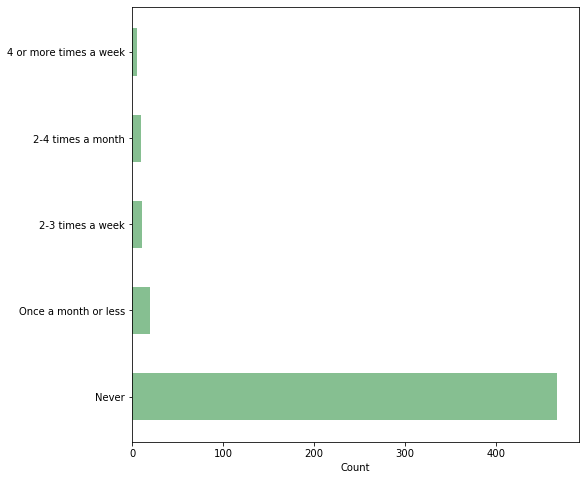}
  \caption{Q21:  In January 2020, approximately how often did you use marijuana/cannabis (recreational or medical)?}
  \label{fig:sub2}
\end{subfigure}

\begin{subfigure}{.33\textwidth}
  \centering
  \includegraphics[width=1\linewidth]{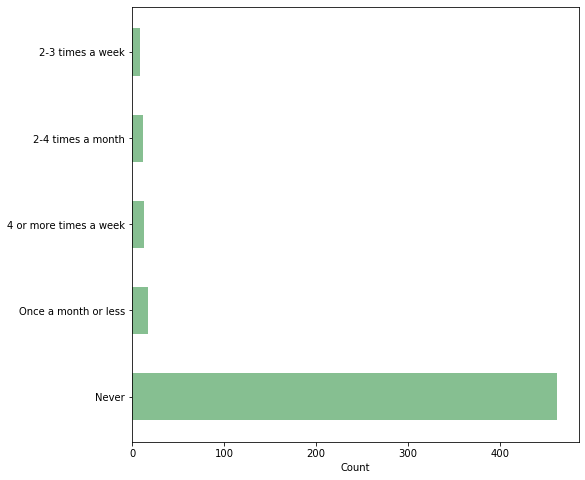}
  \caption{Q22:  In the last month, approximately how often did you use marijuana/cannabis (recreational or medical)?}
  \label{fig:sub1}
\end{subfigure}%
\begin{subfigure}{.33\textwidth}
  \centering
  \includegraphics[width=1\linewidth]{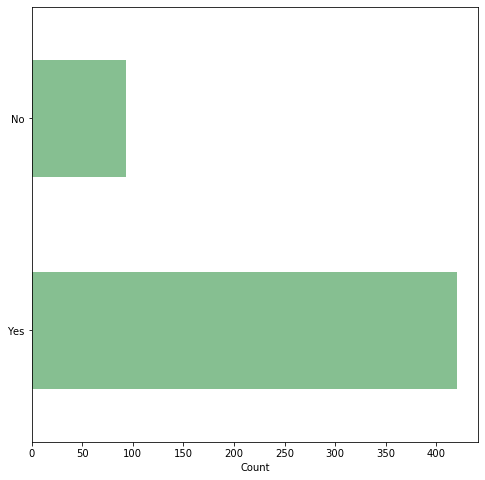}
  \caption{Q23:  Has the amount of news you are consuming increased since the end of Feb, 2020?}
  \label{fig:sub2}
\end{subfigure}
\begin{subfigure}{.33\textwidth}
  \centering
  \includegraphics[width=1\linewidth]{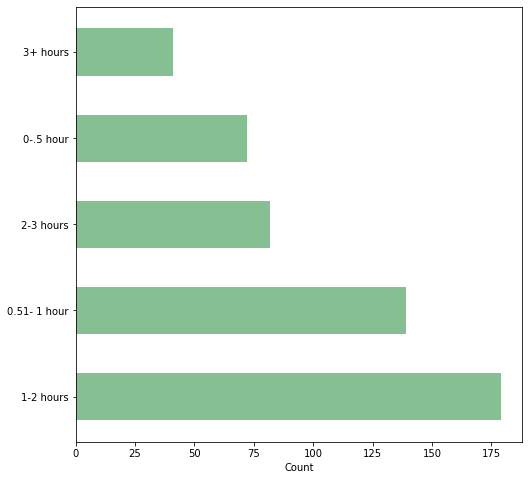}
  \caption{Q24:  How many hours of COVID-19 related news or social media are you consuming on average per day?}
  \label{fig:sub2}
\end{subfigure}
\caption{Frequency bar charts after the preprocessing phase}
\label{fig:test}
\end{figure}

\begin{figure}
\centering
\begin{subfigure}{.33\textwidth}
  \centering
  \includegraphics[width=1\linewidth]{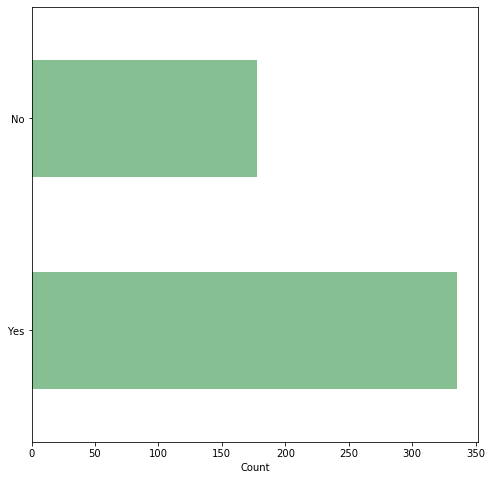}
  \caption{Q25: Have you had more “screen time” (e.g. use of smartphone, tablet, etc.) around bedtime?}
  \label{fig:sub1}
\end{subfigure}%
\begin{subfigure}{.33\textwidth}
  \centering
  \includegraphics[width=1\linewidth]{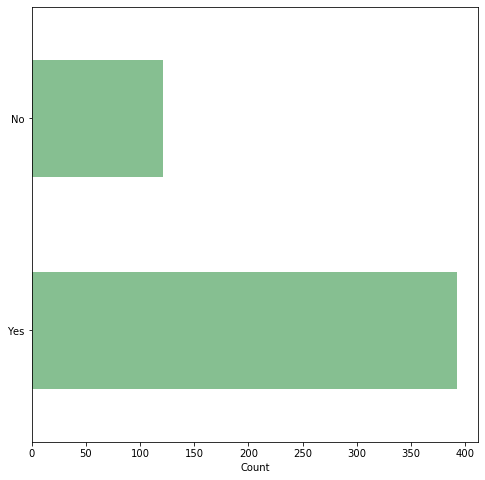}
  \caption{Q26: Have you changed your movement/exercise? }
  \label{fig:sub2}
\end{subfigure}
\begin{subfigure}{.33\textwidth}
  \centering
  \includegraphics[width=1\linewidth]{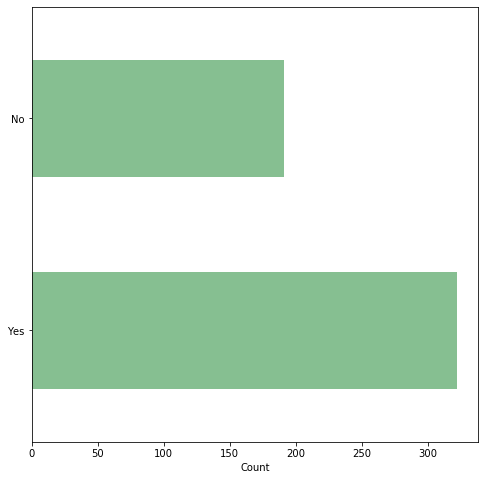}
  \caption{Q27: Has the quality of your diet changed?}
  \label{fig:sub2}
\end{subfigure}

\begin{subfigure}{.33\textwidth}
  \centering
  \includegraphics[width=1\linewidth]{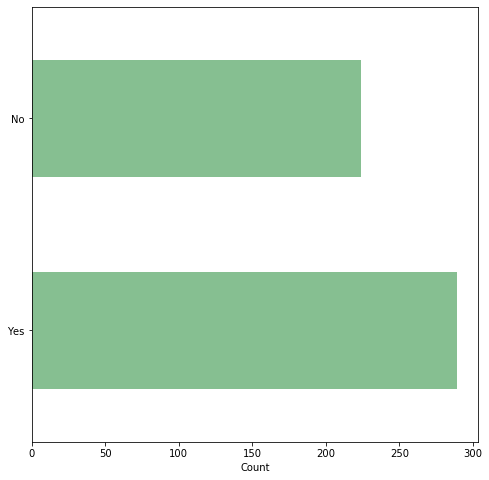}
  \caption{Q28: Has the amount of food you have been eating per day changed?}
  \label{fig:sub1}
\end{subfigure}%
\begin{subfigure}{.33\textwidth}
  \centering
  \includegraphics[width=1\linewidth]{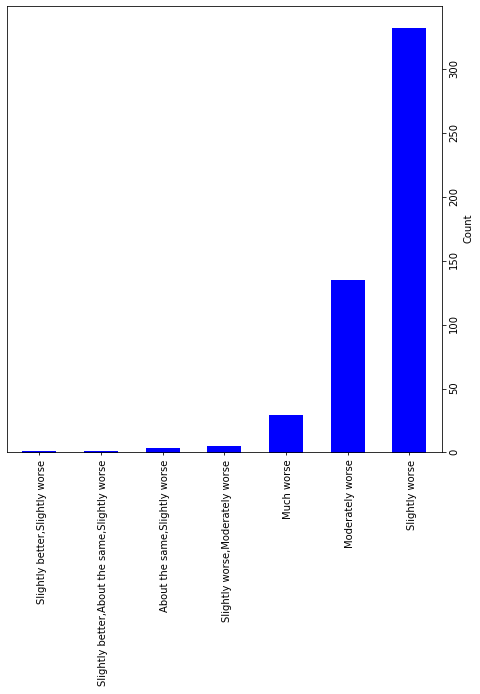}
  \caption{Q29 (a): Please tell us how your mood has changed.  My mood has been:}
  \label{fig:sub2}
\end{subfigure}
\begin{subfigure}{.33\textwidth}
  \centering
  \includegraphics[width=1\linewidth]{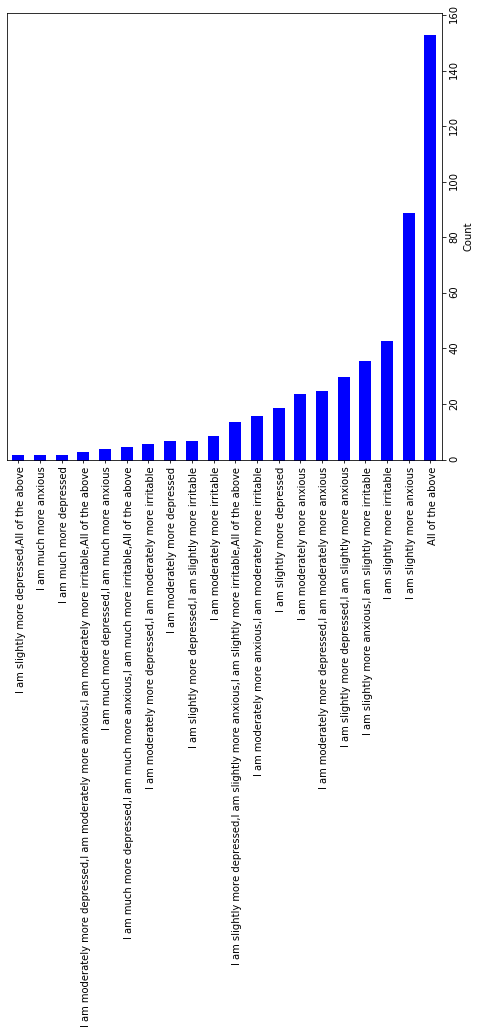}
  \caption{Q29 (b): Please tell us how your mood has worsened  }
  \label{fig:sub1}
\end{subfigure}%
\caption{Frequency bar charts after the preprocessing phase}
\label{fig:test}
\end{figure}